\documentclass[journal]{IEEEtran}
\usepackage{amsmath,amsfonts}
\usepackage{algorithmic}
\usepackage{array}
\usepackage[caption=false,font=normalsize,labelfont=sf,textfont=sf]{subfig}
\usepackage{textcomp}
\usepackage{stfloats}
\usepackage{url}
\usepackage{verbatim}
\usepackage{graphicx}
\def\BibTeX{{\rm B\kern-.05em{\sc i\kern-.025em b}\kern-.08em
    T\kern-.1667em\lower.7ex\hbox{E}\kern-.125emX}}
\usepackage{balance}

\usepackage{paralist}

\usepackage[dvipsnames]{xcolor}

\usepackage[colorinlistoftodos]{todonotes}
\begin{document}
\title{Grand Challenges in the\\  Verification of Autonomous Systems}
\author{Kevin Leahy~\IEEEmembership{Member, IEEE},  Hamid Asgari~\IEEEmembership{Senior Member, IEEE}, Louise A. Dennis, Martin S. Feather, Michael Fisher~\IEEEmembership{Member, IEEE}, Javier Ibanez-Guzman~\IEEEmembership{Life Member, IEEE}, Brian Logan, Joanna I. Olszewska~\IEEEmembership{Senior Member, IEEE}, and Signe Redfield~\IEEEmembership{Member, IEEE}
\thanks{A portion of this research was carried out at the Jet Propulsion Laboratory, California Institute of Technology, under a contract with the National Aeronautics and Space Administration (80NM0018D0004). This work was supported by the Royal Academy of Engineering through its Chairs in Emerging Technologies scheme and by UKRI through research grants EP/X02489X, EP/W01081X, and EP/V026801.}
\thanks{K. Leahy is with the Robotics Engineering Department, Worcester Polytechnic Institute, Worcester, MA, USA.} \thanks{H. Asgari is with Complex and Autonomous Systems Research Group, Thales UK Research, Reading, UK.} \thanks{ M. S. Feather is with Jet Propulsion Laboratory, California Institute of Technology, Pasadena, CA, USA.} \thanks{L. A. Dennis and M. Fisher are with the Department of Computer Science, University of Manchester, Manchester, UK.} \thanks{J. Ibanez-Guzman is with Renault, Guyancourt, France.} \thanks{B. Logan is with the Department of Computing Science, University of Aberdeen, Aberdeen, UK and the Department of Information and Computing Sciences, Utrecht University, Utrecht, The Netherlands.} \thanks{J. I. Olszewska is with the School of Computing and Engineering, University of the West of Scotland, Glasgow, UK.} \thanks{S. Redfield is with the U.S. Naval Research Laboratory, Washington, D.C., USA.}}

\markboth{Journal of \LaTeX\ Class Files,~Vol.~18, No.~9, September~2023}%
{How to Use the IEEEtran \LaTeX \ Templates}

\maketitle

\begin{abstract}
Autonomous systems use independent decision-making with only limited human intervention to accomplish goals in complex and unpredictable environments. As the autonomy technologies that underpin them continue to advance, these systems will find their way into an increasing number of applications in an ever wider range of settings. If we are to deploy them to perform safety-critical or mission-critical roles, it is imperative that we have justified confidence in their safe and correct operation. Verification is the process by which such confidence is established. However, autonomous systems pose challenges to existing verification practices. 
This paper highlights viewpoints of the Roadmap Working Group of the IEEE Robotics and Automation Society Technical Committee for Verification of Autonomous
Systems,
identifying these grand challenges, and providing a vision for future research efforts that will be needed to address them.
\end{abstract}

\begin{IEEEkeywords}
Autonomous systems, system verification, safety, reliability
\end{IEEEkeywords}

\section{Introduction}

An autonomous system is told what to achieve and the system itself decides how to achieve it.  Autonomous systems are thus needed in situations where circumstances cannot be predicted in advance and there are too many variations to program the response to each one. They are increasingly being deployed in a broad spectrum of environments and for a wide range of purposes; for example, autonomous vehicles are being used on land, in the air, on or under the sea, and in space, for applications including transportation, exploration, surveying and extraction.
As autonomous systems proliferate and technology advances, these systems will operate more independently, including in safety-critical, mission-critical, and remote applications. In all of these settings, it is imperative that systems operate safely, correctly, and reliably. 
Verification is an important process for ensuring safe and correct behavior~\cite{araujo2023}. 

Informally, verification is the process of determining whether a system meets a set of requirements.
Verification is widely used in industry to establish properties of hardware and software systems. 
However, the verification of autonomous systems brings new challenges. 
The wide variety of situations in which they are deployed preclude exhaustive testing as the primary means to verify their correctness. 
The nature of their programming can differ substantially from that of automatic systems – they may include planners and schedulers that formulate on-the-fly the actions to achieve their goals, and they may maintain and reason about internal models of themselves and of the environment in which they operate based on observations.
When their actions may impact humans, other systems or the environment, they will be expected to follow societal norms and behave both legally and ethically – phenomena we naturally associate with humans but have little experience programming in software, let alone verifying. 
In addition, autonomous systems may be expected to adapt to their environment by learning to better understand that environment or to find better ways of achieving their goals, in which case verification must address not only the systems when initially deployed, but also that as they evolve over time they will continue to behave as would be intended.

While today's autonomous systems work in relatively narrow domains with fairly limited independence, the field is advancing rapidly, and it is clear that future autonomous systems will have greater decision-making power, across a wider array of domains, with a greater ability to directly impact the world than current software systems. The next stage of autonomous systems will represent an inflection point in autonomous behaviors, and with these increased capabilities come a greater degree of responsibility. 
Therefore, there is a need to prepare our verification practices to meet this change, especially at the \emph{behavioral} level. It is imperative that we are capable of verifying these systems before they reach our day-to-day lives, not after.

In this paper, we lay out a vision of future research efforts to verify the behavior of autonomous systems in the real world.
It represents a consensus viewpoint among members of the Roadmap Working Group of the IEEE Robotics and Automation Society Technical Committee for Verification of Autonomous Systems\footnote{\url{https://www.ieee-ras.org/verification-of-autonomous-systems}}, which comprises representatives from industry, academia, and government.
We believe that the vision we present complements previous visions and road maps for verification or assurance of autonomous systems. Typically, these tend to be either domain-specific, such as for air vehicles~\cite{brat2023autonomy} or the internet of things~\cite{ratasich2019}, or otherwise very broad~\cite{topcu2020,VAMS23}, capturing many aspects of autonomy, including education and governance. Here, we focus specifically on verification, and the need for new approaches and tools for assuring safe and reliable autonomous behavior.
While validation, evidence collection, and certification are also important and challenging topics for the deployment of autonomous systems, we deliberately limit the scope of this work to verification.
We note that there exist standards, such as IEEE 1012~\cite{ieee1012}, that address verification and validation of a broad range of systems, software, and hardware. These standards provide guidance on life cycle processes, risk assessment, and the steps involved in verification and validation. By contrast, our work focuses on issues related to autonomous systems that provide challenges for verification.
Finally, this document also does not focus specifically on the issue of artificial intelligence (AI). We view AI as an important component of many autonomous systems, but somewhat orthogonal to the issue of verifying autonomous systems themselves. Verification of AI is important in general, including for verification of autonomous systems. However, the issue of AI verification has been addressed directly by others, e.g.,~\cite{seshia2022}.

\section{The Verification Process}
\begin{figure}
    \centering
    \includegraphics[width=\columnwidth]{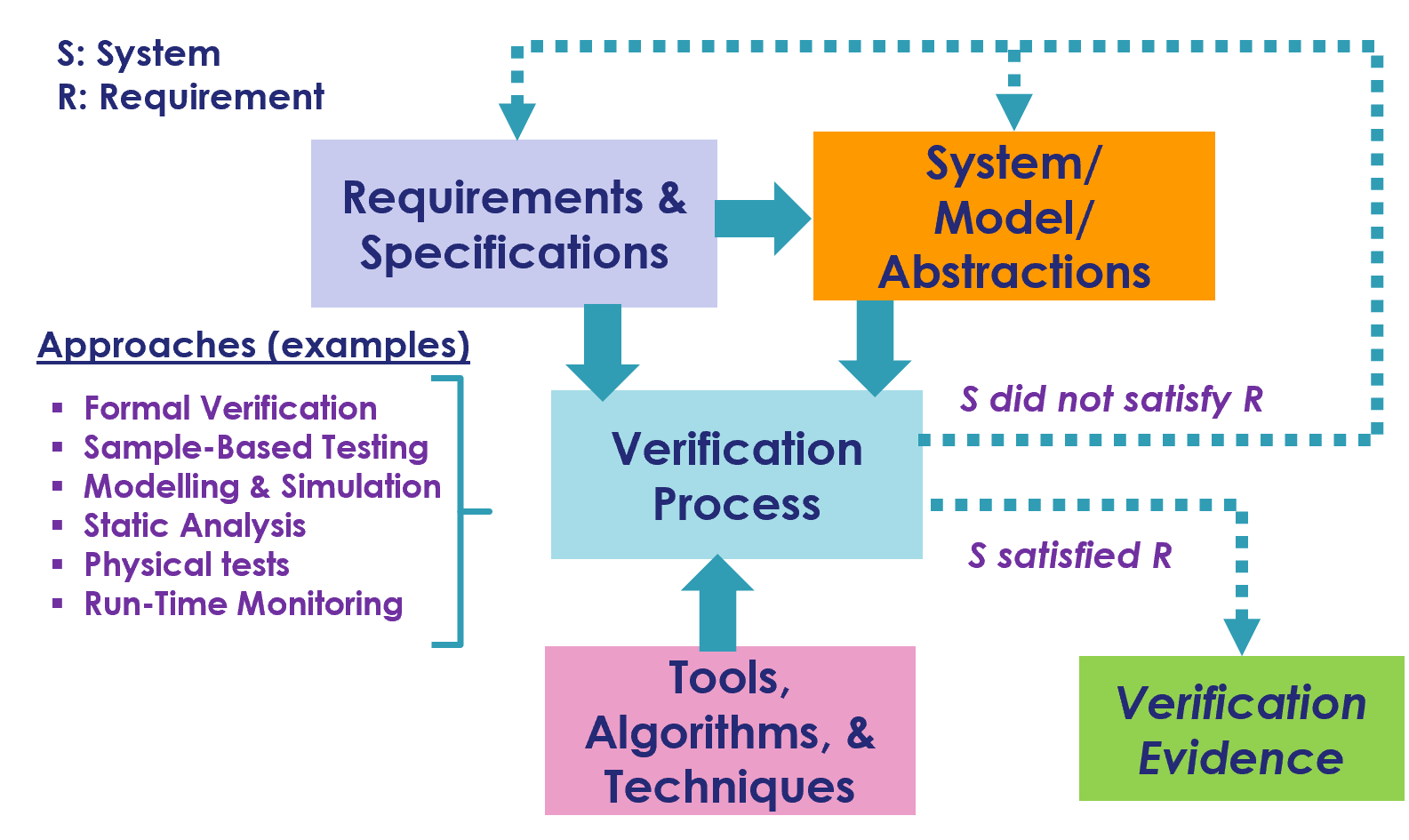}
    \caption{The verification process directs the generation of evidence that will demonstrate the system meets its requirements. It specifies the approach, or combination of approaches, to be followed. These in turn use tools, algorithms, and techniques to generate verification evidence.}
    \label{fig:verification_process}
\end{figure}

While there are many approaches to verification, 
the key elements of requirements and specifications, models and abstractions, and tools and algorithms are common to all
(see Figure~\ref{fig:verification_process}).
The verification process seeks to establish whether a system satisfies a set of requirements. In some cases, this can be done using precise mathematical procedures. For example, we may be able to use logical proofs to confirm that the requirements (specified in an appropriate mathematical or logical notation) are satisfied. However, for many complex systems, such formal verification may not be possible, and we must utilize other techniques such as testing, or formal verification of a \emph{model} of the system in question, rather than of the system itself. For example, we may \emph{abstract} the system into a model, ensuring that the key characteristics of the system relevant to the requirements are preserved. Then, we can formally verify the requirements with respect to the abstraction. Another approach to handling complexity is to verify the requirements by carrying out a large number of (synthetic) tests with respect to a \emph{simulation} of the system in its operating environment. This approach can be particularly useful when the actions of the system are potentially harmful in the real world.

\section{What Makes Autonomy Difficult?}
There is no universal agreement about what constitutes ``autonomy". However, there are several important characteristics of autonomous systems around which there is some consensus. Relevant to this work are: \begin{inparaenum}
    \item independent decision-making to accomplish goals;
    \item limited human intervention; and
    \item operation in complex and unpredictable environments.
\end{inparaenum}
A system need not be particularly intelligent to be autonomous. For example, the work of Braitenberg~\cite{braitenberg1986} showed how seemingly complex behavior can emerge from the interaction of relatively simple systems and their environment. Rather, it is a system's ability to make its own decisions in a complex environment that is the salient aspect of autonomy. 
It is also important distinguish between \emph{automation} and \emph{autonomy}. While a robot in an assembly plant may operate independently (and even intelligently), its environment and the conditions it may encounter are strictly controlled, allowing all aspects of its behavior to be predicted, including the decisions it makes. We therefore consider such systems to be automated rather than autonomous. 

These characteristics of autonomous systems give rise to two main interrelated challenges which create difficulties for the verification process: uncertainty and context.

\paragraph{Uncertainty} 
Uncertainty, particularly epistemic uncertainty (i.e., uncertainty due to limited knowledge), presents a major challenge for the verification of autonomous systems. In a real-world environment, there are many ``unknown unknowns" that are not anticipated or understood at design and validation time. This is particularly true where systems are deployed in settings that differ from the environment for which they were designed, where the environment changes over time, or where the autonomous system itself causes changes to the environment. As noted above, requirements are often verified against \emph{models} of the system and the environment in which it operates. If there are aspects of the environment that are unknown, modeling those aspects is inherently impossible, forcing a designer to rely on expectations about what could happen. Recognizing and modeling the occurrence of unanticipated scenarios is therefore a key challenge, as it relies on a model of ``unknown unknowns", 
and defining and verifying appropriate behavior in such cases is also a challenge.

\paragraph{Context} 
Many autonomous systems that interact with humans and/or other agents, for example, in autonomous driving scenarios, domestic and healthcare robotics, or even non-embodied agents such as high-frequency trading, will have legal, ethical, and social responsibilities. We have always required our computational systems to be legal and ethical, but traditionally these properties have been considered as non-functional requirements handled at design time, and not explicitly included as part of verification processes.  
As systems become more autonomous, the satisfaction of these non-functional requirements depends more upon the autonomous decision-making and functional behaviour of the system.  We therefore require methodologies for translating such non-functional requirements into functional ones. These then can be handled by verification processes targeted at assessing functional behaviour, using methods such as testing, model-checking and theorem proving-based approaches. This need to make non-functional requirements functional applies not only to legal, ethical and social requirements but also to other widely recognised non-functional requirements such as transparency, fault tolerance, resilience, and robustness.

In general, the satisfaction of a functional property derived from a non-functional requirement depends upon the circumstances in which the system makes a decision. Therefore, we potentially need to widen our view of ``environment'' to encompass less tangible aspects such as the ethical, legal and social factors that may be in play.  Similarly, addressing fault tolerance, resilience, and robustness may require extending our concept of environment to handle the epistemic uncertainties discussed in the previous section. Broadly, we can categorize this as the \emph{context} within which the system is operating. 
Techniques for describing and reasoning about the physical or computational environment may be very different to those for describing and reasoning about the wider context. After all, 
judging whether the system has correctly diagnosed that some healthcare intervention is required based on observations is different to deciding what the best intervention may be, how this should be described or explained to a patient, whether their participation should be encouraged, and whether others (and if so whom) should be recruited to help convey information and encourage consent.

In many, if not most, cases, an autonomous system will need to maintain its own abstraction of the environment/context in order to guide its decisions. 
It is not clear how to verify that such an abstraction is correct, or matches the environment/context it is meant to model. Further, the effects of the abstraction on decision-making, and therefore on system behavior, are likewise not well understood. In the worst case, the designer and autonomous system may only have a minimal representation of the environment, but we would hope the autonomous system may still make the ``correct" decision. Defining what it means to be correct in such situations is itself a challenge.

\section{Recommendations}\label{sec:recommendations}
The characteristics of autonomous systems described above require new approaches to verification. In this section, we describe our recommendations for future research directions to address these challenges. 

\subsection{Requirements}
The formulation of precise, consistent and complete requirements for an autonomous system is a necessary prerequisite to its verification. 
When humans are the decision makers, we expect them to adopt a ``reasonable” course of action. 
However, what is obviously reasonable to humans in a given situation is very difficult to formulate as a set of requirements for an autonomous system. In addition, engineers are used to prescribing functional requirements in relatively narrow operating domains. An autonomous system operating in the real world will interact with many more entities—human or otherwise—effectively involving many new stakeholders, from users to bystanders to regulators, and introducing new issues of behavioral, legal, and ethical requirements. Specifying ``reasonable'' behavior in this sense is both critical and difficult.

Concretely, there is a need to develop processes and standards for specifying requirements that cover ethical, legal and social expectations. These in turn will need to be translated to specifications that can be used for verification purposes. 
While there is some work being done on, for example, ethical~\cite{dennis2016formal} or responsible behavior~\cite{shalev2017formal}, there is a need for a more unified approach. Such a unified approach will help to identify classes of properties that cannot currently be expressed by existing specification languages. 
There may be properties that are not decidable, or that require new verification techniques. 

Requirements expressed in terms of the set of behaviors of the system in its possible environments need to be refined into functional requirements as a prerequisite to their verification. For example, the level of safety required of a system may be expressed as “the frequency of accidents shall be less than 1 per million hours of operation when driven in daylight conditions on major highways”. Refinement is necessary for verification purposes, as gathering sufficient real-world data to have statistical results would mean lengthy fielding of unverified systems while that data is being accumulated. When engineering a conventional system, such safety properties are turned into functional requirements by deriving reliability properties required of the elements of the system --- its sensors, actuators, and control system --- such that in combination they will achieve the safety requirement. This approach will need to evolve to address the novel forms of reasoning that future autonomous systems will employ when interpreting sensory inputs,  planning future actions, and executing control decisions. Furthermore, 
research is needed 
to verify the correctness and completeness of those refinements, and thereafter how then to verify the functional requirements themselves on individual executions of the system.

\subsection{Semantics}
Verifying that functional requirements on a system’s behavior guarantee a non-functional requirement on the system as a whole will typically involve different levels of detail, and different areas of concern, and require advances in reasoning involving compositional and heterogeneous semantics.
Generally, we think of desirable behavior in terms of interactions in the environment, e.g., ``if you see a pedestrian, stop at least ten feet away". Such requirements necessitate reasoning about how a system \emph{perceives} a pedestrian at a sensor or signal level, what it \emph{decides} to do upon recognizing a pedestrian, and how it \emph{executes} the resulting actions. Each of these components has different semantics, and the requirements we specify for those components likewise have different semantics. The situation is further complicated if the requirements are more abstract, in terms of ethical or legal issues. Verification at any one of these levels of abstraction may be feasible, but crossing these semantic boundaries is non-trivial, and needs attention if we wish to make claims on system-level behavior. 

\subsection{Operating envelopes}
Autonomous systems are expected to function independently in complex environments. Inevitably, something will diverge from what was expected, be it the environment, the autonomous system's behavior, or something else. 
Here, we present recommendations for dealing with unanticipated factors, and how we may address them for verification.

At the behavioral level, some tools exist for expressing desired behavior and even synthesizing designs to satisfy that behavior. However, there is less work for specifying reasonable behavior when the system is operating in circumstances in which desirable behavior is no longer achievable. Runtime verification and monitors can identify when a system is no longer meeting requirements, but they cannot decide what to do about it. For instance, a monitor can detect that the braking behaviour of a vehicle no longer slows it down as rapidly as required but the monitor alone cannot then decide between braking earlier in future, coming to a halt altogether, only driving in certain less hazardous conditions, etc. Identifying and verifying the correct fallback behavior is non-trivial.

There is a need to develop methods for detecting when autonomous systems' models are no longer reflective of the object or process that they are meant to represent. This is related to the notions of out-of-distribution (OOD) detection in machine learning, as well as runtime verification and monitoring. However, neither of these approaches are designed to identify when a model is no longer reflective of an autonomous system or its environment. 
To deploy an autonomous system, methods must be developed to distinguish between the system being out of its operating envelope and when its operating envelope is not properly represented by the model the autonomous system is using.

A useful notion from the automotive domain is that of the Operational Design Domain (ODD)~\cite{J3016_202104}, defined there as the ``operating conditions under which a given driving automation system or feature thereof is specifically designed to function, including, but not limited to, environmental, geographical, and time-of-day restrictions, and/or the requisite presence or absence of certain traffic or roadway characteristics."
This is a useful framework for reasoning about automotive behavior, where vehicles operate within the relatively structured environment of roadways. Autonomous systems operating in other environments, such as inside people's homes, will need to deal with situations and scenarios that vary in different ways. Formalizing and abstracting the notion of ODD for a broader range of autonomous systems will be helpful for identifying the conditions in which we expect them to operate correctly, and those in which fallback behaviors or other measures are needed.

As noted earlier, an autonomous system may operate in environments that differ from those for which it was initially verified, and an autonomous system that includes a learning-enabled component may adjust its behavior based on its experiences. In such cases, the autonomous system itself should determine whether, and if so, the extent to which those differences or changes have weakened or invalidated the initial verification of the autonomous system, and respond accordingly, for example, by alerting its users to its reduced reliability. This requires an autonomous system to have knowledge of, and the ability to reason over, its own verification. A step in this direction is the “Dynamic Assurance Cases” concept proposed in \cite{asaadi2020}, which comprise various system information, including requirements and their traceability to other artifacts that may change dynamically after deployment. Ambitiously, we may wish for an autonomous system to be self-reverifying, i.e., take actions to autonomously update its own verification (e.g., itself run more simulations or tests to assess its revised suitability). Verifying the autonomous reverifier then becomes a challenge!

\subsection{Novel verification techniques}

The nature of autonomous systems programming can differ substantially from that of automatic systems. As noted above, autonomous systems may include diagnostic subsystems that determine the situation from observations, planners and schedulers that formulate on-the-fly the actions to achieve their goals, and inference engines that draw probabilistic conclusions from rich sources of data. They may maintain and reason about internal models of themselves and of the environment in which they operate. These differences must be accounted for in verification techniques.

Techniques are needed for verification of the kinds of software used by autonomy, notably: \begin{inparaenum}
\item the planners and schedulers that reason about the goals the autonomous system is to achieve;
\item the software to not only detect anomalies (in the system itself, or of deviations of the environment from the assumptions made when the system was developed) but also to identify their causes (if necessary directing additional actions specifically to aid in such identification); and finally \item to develop appropriate responses\end{inparaenum}. When, as is often the case, autonomy software relies upon models of its own system and the environment in which it operates, verification must extend to the models as well as the software that uses them. Traditional practices --- such as measuring code coverage to assess the thoroughness of testing,  metrics to estimate code complexity, and coding standards that guide the development of software to avoid common pitfalls --- all need equivalent practices to deal with models. 

Finally, decision-making, central to many autonomous systems, requires further verification research. Although the verification of relatively simple decisions, undertaken in well understood systems, has been a popular topic within the agent-based systems community, unsolved  issues relating to verification remain. (Note that we do not consider the explanations of these decisions, a topic that is important in itself.)

The first issue concerns the varieties of decisions being made. Both specifying and verifying the range and complexity of decisions that the autonomous system might need remains a difficulty. 
As noted above, work is needed to provide coherent and consistent approaches to the specification of a wide variety of decisions. 
Complexity in these specifications might then necessitate new verification approaches and new computational issues. A further issue is that validation should ideally be incorporated with verification and so heterogeneous approaches to verification and validation may well be needed in this area.

A second issue around decision-making and its verification concerns the range of uncertainties that come together in this area. There are obvious uncertainties about the system's environment, but in combination with our own uncertainties about this knowledge (the ``unknown unknowns'') a dauntingly complex context can result, within which decision-making should take place. A further factor is the increasing use of often quite opaque AI technologies within the autonomous system, and verifying behaviour and decisions in such cases remains an unsolved problem. The fact that these systems might be huge, that any assumptions about the distribution of data will very likely be wrong, and that the components are likely to be highly dynamic in nature, all make the verification aspects difficult and, as yet, unsolved.

\subsection{A wider view of verification}
Motivating our study is the recognition that the nature of autonomous systems poses challenges to existing verification practices, and yet applications of such systems and the need for their verification are proliferating in ever more settings and for a growing variety of purposes. Correspondingly, verification of their correct operation will be of interest to a broadening range of parties. These parties include those developing, deploying and maintaining such systems; the responsible investors, regulators, certification agencies, and insurers; and of course those directly affected by the actions or inaction of autonomous systems: users and bystanders. This diverse mix will soon also include autonomous systems themselves, deciding whether and how to interact with one another. Thus, the way in which the verification processes and results are presented and communicated cannot be one-size-fits all, but will need to be tailored to the recipient of that information. For example, a user may want insight into the system's behavior, without requiring detailed information better suited to a developer. The emergence of safety cases (and their generalization to assurance cases) marked a transition from claiming verification of a system through adherence to all-purpose prescriptive standards, to instead marshalling the evidence as a system-specific argument. We see the need for a follow-on transition in which the assurance argument can be expressed in multiple forms, appropriate to a specific audience.

\section{Timeline and Conclusion}
\begin{figure}
    \centering
    \includegraphics[width=\columnwidth]{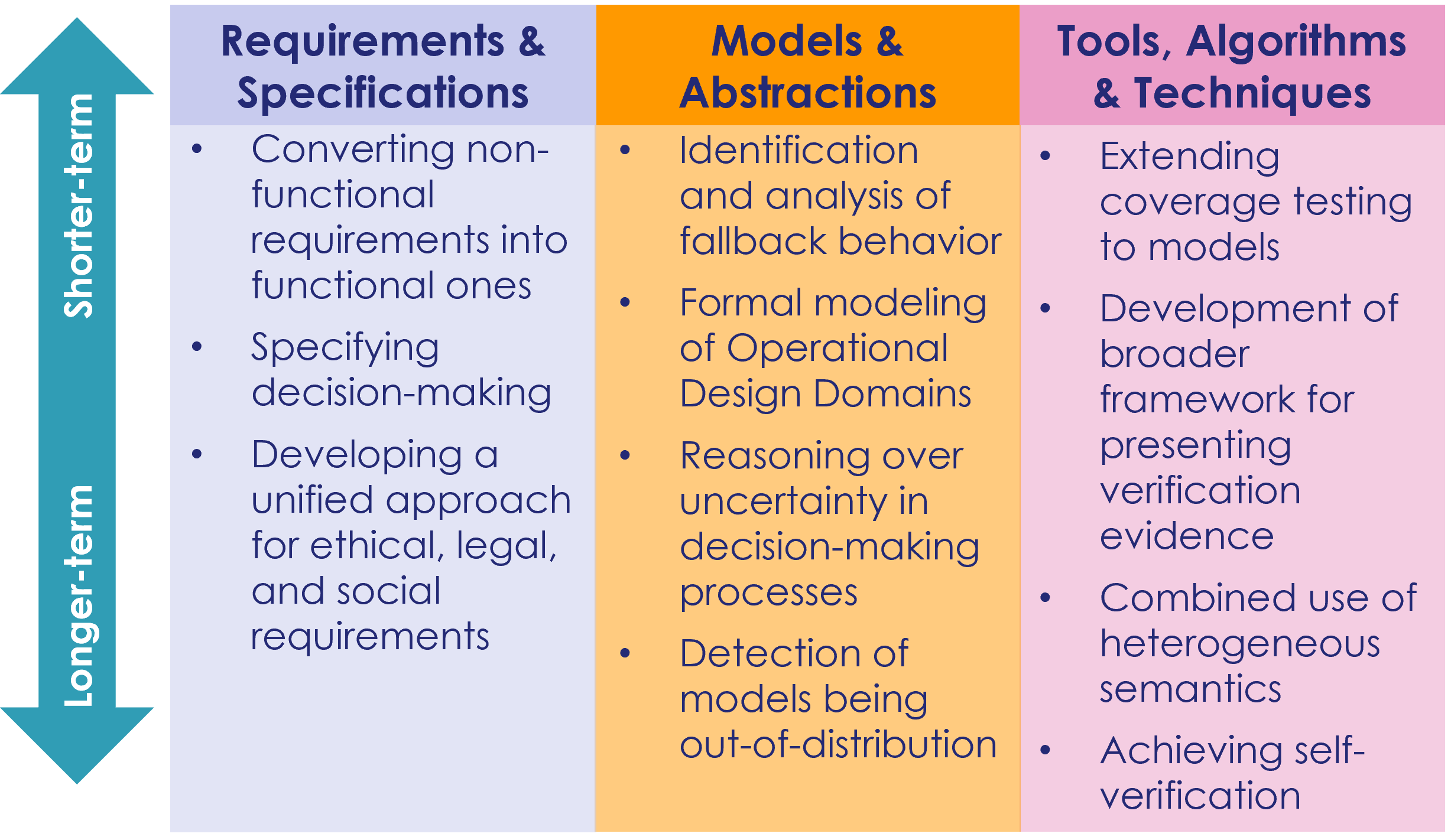}
    \caption{The recommendations identified in Sec.~\ref{sec:recommendations}, categorized according to the components of verification identified in Fig.~\ref{fig:verification_process}, and ordered by how soon they are likely to be able to be put into practice. Items appearing closer to the top are shorter-term objectives than those lower down, but no specific timescale is implied.}
    \label{fig:roadmap}
\end{figure}

We summarize our recommendations from the preceding sections in the timeline in Fig.~\ref{fig:roadmap}, grouped according to the verification process depicted in Fig.~\ref{fig:verification_process}. They are presented vertically from shorter-term to longer-term. We do not predict how long each development will take, because this is likely to vary widely between different applications and industries. Instead, we have presented them relative to the order in which broadly usable solutions are likely to appear. 

As autonomous system development continues apace, verification of those systems must keep up. Just as there are many novel technologies and theories required to create autonomous systems, so too is there a great need for novel technologies and theories to support their verification. Here, we have attempted to define specifically what challenges autonomous systems present from a verification perspective and what potential avenues there are for addressing those challenges. We are optimistic that the scientific and engineering community will overcome these challenges, leading to autonomous systems that are not just capable, but safe and reliable as well.

\section{Acknowledgements}
\noindent The authors wish to thank Dejanira Araiza Illan and John R. Budenske for their thoughtful comments and input in the writing of this manuscript.

\bibliographystyle{ieeetr}
\bibliography{thebib}

\end{document}